\documentclass[10pt]{article}
\usepackage{fullpage}
\usepackage[utf8]{inputenc}
\usepackage{amsmath}
\usepackage{amsfonts}
\usepackage{amssymb}

\usepackage{mathrsfs}
\usepackage{color,soul}
\usepackage{pdflscape}
\usepackage{booktabs}
\usepackage{rotating}
\usepackage{makecell}
\usepackage[english]{babel}

\usepackage{blindtext}

\usepackage{adjustbox}

\usepackage{booktabs}
\usepackage{float}
\usepackage{multirow}
\usepackage{fancybox,graphicx}
\usepackage{subfig}
\usepackage{caption}
\usepackage{subcaption}
\usepackage{color}
\usepackage{xcolor}
\usepackage{authblk}
\usepackage[colorlinks]{hyperref}

\usepackage{hyperref}

\newcommand{\doi}[1]{{doi:~\href{https://doi.org/#1}{\nolinkurl{#1}}}\rmFullStop}

\newcommand*{\rmFullStop}{\rmifnextchar{.}{}{}}

\makeatletter
\newcommand{\rmifnextchar}[3]{%
  \begingroup
  \ltx@LocToksA{\endgroup#2}%
  \ltx@LocToksB{\endgroup#3}%
  \ltx@ifnextchar{#1}{%
    \def\next{\the\ltx@LocToksA}%
    \afterassignment\next
    \let\scratch= %
  }{%
    \the\ltx@LocToksB
  }%
}
\makeatother 

\usepackage[titletoc,title]{appendix}
\usepackage{cite}
\usepackage{devanagari}
\usepackage[T1]{fontenc}
\usepackage{microtype}
\usepackage{mathtools}

\title{CorIL: Towards Enriching Indian Language to Indian Language Parallel Corpora and Machine Translation Systems} 
 
\author[1]{Soham Bhattacharjee}
\author[2]{Mukund K Roy}
\author[3]{Yathish Poojary}
\author[4]{Bhargav Dave}
\author[5]{Mihir Raj}
\author[6]{Vandan Mujadia}
\author[1]{Baban Gain}
\author[11]{Pruthwik Mishra}
\author[6]{Arafat Ahsan}
\author[6]{Parameswari Krishnamurthy}
\author[3]{Ashwath Rao}
\author[7]{Gurpreet Singh Josan}
\author[8]{Preeti Dubey}
\author[9]{Aadil Amin Kak}
\author[10]{Anna Rao Kulkarni}
\author[3]{Narendra VG}
\author[2]{Sunita Arora}
\author[5]{Rakesh Balbantray}
\author[4]{Prasenjit Majumdar}
\author[2]{Karunesh K Arora}
\author[1]{Asif Ekbal}
\author[6]{Dipti Mishra Sharma}

\affil[1]{Department of Computer Science and Engineering, Indian Institute of Technology Patna}
\affil[2]{SNLP Lab, CDAC Noida}
\affil[3]{Department of Computer Science and Engineering, Manipal Institute of Technology}
\affil[4]{Department of Computer Science and Engineering, Dhirubhai Ambani University, Gandhinagar}
\affil[5]{Department of CSE, IIIT Bhubaneshwar}
\affil[6]{LTRC, IIIT Hyderabad}
\affil[7]{Department of CSE, Punjabi University}
\affil[8]{Department of CSE, Govt. College for Women Jammu}
\affil[9]{Department of Linguistics, University of Kashmir}
\affil[10]{VLSI Design Group, CDAC Bangalore}
\affil[11]{Department of AI, SVNIT, Surat}

\begin{document}

    \maketitle

\begin{abstract}



\color{black}

India's linguistic landscape is one of the most diverse in the world, comprising over 120 major languages and approximately 1,600 additional languages, with 22 officially recognized as scheduled languages in the Indian Constitution \footnote{\url{https://en.wikipedia.org/wiki/Languages_of_India}}. Despite recent progress in multilingual neural machine translation (NMT), high-quality parallel corpora for Indian languages remain scarce, especially across varied domains. In this paper, we introduce a large-scale, high-quality annotated parallel corpus covering 11 of these languages—English, Telugu, Hindi, Punjabi, Odia, Kashmiri, Sindhi, Dogri, Kannada, Urdu, and Gujarati—comprising a total of 772,000 bi-text sentence pairs. The dataset is carefully curated and systematically categorized into three key domains: Government, Health, and General, to enable domain-aware machine translation research and facilitate effective domain adaptation.

To demonstrate the utility of CorIL and establish strong benchmarks for future research, we fine-tune and evaluate several state-of-the-art NMT models, including IndicTrans2, NLLB, and BhashaVerse. Our analysis reveals important performance trends and highlights the corpus's value in probing model capabilities. For instance, the results show distinct performance patterns based on language script, with massively multilingual models showing an advantage on Perso-Arabic scripts (Urdu, Sindhi) while other models excel on Indic scripts. This paper provides a detailed domain-wise performance analysis, offering insights into domain sensitivity and cross-script transfer learning. By publicly releasing CorIL, we aim to significantly improve the availability of high-quality training data for Indian languages and provide a valuable resource for the machine translation research community. 


\end{abstract}

\section{Introduction}
India is one of the most linguistically diverse countries in the world.  While English and Hindi functions as a common medium for formal communication, particularly in sectors such as business, education, governance, and the judiciary, the multilingual nature of the country underscores the critical need for effective language translation systems.

Language translation in India is not merely a matter of convenience but a necessity for fostering social inclusion, ensuring equitable access to information, and maintaining national cohesion. For instance, the successful dissemination of government policies, public welfare schemes, and legal rights relies heavily on the availability of content in regional languages. Furthermore, the country grapples with a significant shortage of medical professionals, particularly in rural areas and community health centers. In such contexts, machine translation can serve as a vital tool to bridge the communication gap between healthcare providers and patients across linguistic boundaries, thereby enhancing the reach of medical services.

Moreover, translation facilitates national integration by enabling seamless communication for individuals migrating or traveling across linguistically distinct regions. In this light, the development of robust machine translation systems tailored to India’s linguistic landscape holds substantial promise in addressing both practical and socio-cultural challenges.
Indian languages present a unique set of challenges for machine translation due to their morphological richness and linguistic diversity. Recent efforts, particularly the development of corpora and models such as Samanantar \cite{samanantar} and IndicTrans2 \cite{gala2023indictrans2}, have marked significant progress in this space. These models were the first to provide translation support for all 22 scheduled Indian languages recognized by the Government of India, collectively spoken by approximately 97\% of the population. However, a persistent limitation in this domain has been the scarcity of high-quality parallel corpora for low-resource languages. The BPCC corpus released alongside IndicTrans2 was constructed through human translation from English to various Indic languages. While valuable, this approach introduces a well-documented issue in low-resource machine translation known as translationese \cite{graham2019translationesemachinetranslationevaluation}, wherein language-specific subtleties and authentic expressions are lost due to the non-native origin of the source text.

To address this, our work, though it caters to only 11 Indian languages, enhances the existing landscape by sourcing texts originally written in low-resource Indian languages and translating them into both Hindi and English. By anchoring our dataset in source-language authenticity, we aim to better preserve linguistic nuances and cultural context. Furthermore, by employing both Hindi and English as pivot languages, we introduce a level of robustness to the data that facilitates the development of future pivot-based machine translation systems.

In this work, we release a new parallel corpus spanning 11 Indian languages and three critical domains, comprising a total of 772,000 sentence pairs. To evaluate the utility of this dataset, we conduct extensive experiments using state-of-the-art models, NLLB-200 \cite{nllbteam2022languageleftbehindscaling}, IndicTrans2 \cite{gala2023indictrans2} and BhashaVerse\cite{mujadia2025bhashaversetranslationecosystem}, analyzing their performance trends after fine-tuning on our released corpus. Additionally, we perform domain-specific fine-tuning to assess the effectiveness of domain adaptation. Beyond parallel data, we also provide a linguistically annotated subset of 133,000 sentences, covering part-of-speech (POS) tagging, named entity recognition (NER), morphological features, and chunking. Notably, approximately 40\% of the collected sentences originate in native low-resource Indian languages, helping to preserve linguistic authenticity. Detailed language-wise and domain-wise data statistics are presented in Table~\ref{tab:splits}. By releasing this corpus to the research community, we aim to democratize access to high-quality translation resources and foster further exploration in the field of Indian machine translation. The corpus is freely available at the following link \url{https://huggingface.co/datasets/HimangY/CoRil-Parallel}


\color{black}
\section{Languages and Domains}

India's linguistic landscape is a testament to its rich cultural heritage. Our dataset encompasses 11 of these languages including, Hindi, Punjabi, Urdu, Telugu, Kannada amongst others.  These languages are written in diverse scripts, many of which trace their origins to the ancient Brahmi script. For instance, Devanagari is used for Hindi, Sanskrit, and Marathi; Bengali script serves both Bengali and Assamese; Gurmukhi is employed for Punjabi; and the Odia script is utilized for the Odia language. Each script is an abugida, wherein consonant-vowel sequences are written as a unit, with inherent vowels modified by diacritics. This structure accommodates the phonetic richness of Indian languages. This section delves into the languages and domains covered by our dataset in greater detail.  
\subsection{Language and Linguistic Properties}


In this work, we started with Hindi and ten Indian languages along with English. The languages involved are Dogri, Hindi, English, Gujarati, Kashmiri, Kannada, Oriya, Punjabi, Telugu, and Urdu. These languages represent three major language families: the Indo-Aryan family includes Dogri, Hindi, Gujarati, Kashmiri, Oriya, Punjabi, and Urdu; the Dravidian family includes Telugu and Kannada; and English, from the Indo-European family, serves as a point of comparison. All the Indian languages considered here follow the Subject-Object-Verb (SOV) word order, whereas English follows the Subject-Verb-Object (SVO) order and uses the Latin script, unlike the Indian languages, which use a variety of Brahmi-derived scripts or the Perso-Arabic script (in the case of Urdu).


\subsubsection*{Sindhi}
Sindhi is an Indo-Aryan language spoken by approximately 30 million people in Pakistan and around 2.5 million in India. It is written in both Perso-Arabic and Devanagari script, with each script adapted to capture Sindhi’s unique phonology. The coexistence of two scripts poses orthographic and standardization challenges. Morphologically, Sindhi exhibits both simple and complex stems formed via derivation, compounding, and reduplication. Verbs are richly inflected for tense, aspect, and mood, and the language employs postpositions and morphological markers for grammatical relations. 

\subsubsection*{Gujarati}

Gujarati is an Indo-Aryan language spoken by over 55 million people, primarily in Gujarat, India, and by diaspora communities worldwide. It uses the Gujarati script, derived from Devanagari but without the top line (\textit{\textbf{śirorekhā}}), and features additional symbols for nasalization. Challenges in this language include dialectal variation and preservation across diaspora settings. Gujarati exhibits three genders and two numbers. Its morphology includes richly inflected nouns, pronouns, and verbs that mark tense, aspect, mood, person, and number. The language also employs postpositions and retains strong phonological ties to Sanskrit.

\subsubsection*{Dogri}

Dogri is an Indo-Aryan language spoken by approximately 2.6 million people, primarily in the Union Territory of Jammu and Kashmir and adjoining regions of North-West India. It was officially recognized as one of India’s national languages in 2003 and has historical roots in Vedic and Shauraseni Prakrit. The language is now written in the Devanagari script, with additional markers like the \textit{avagrah} and apostrophe used to represent specific phonological features such as tone and vowel length. Dogri has 28 segmental and 5 suprasegmental phonemes. Nouns and adjectives are inflected for number, gender, and case, while verbs mark tense, aspect, mood, voice, and person. Despite lexical similarities with Hindi, Urdu, and Punjabi, Dogri preserves distinct phonological and grammatical characteristics.

\subsubsection*{Kannada}
Kannada is a Dravidian language predominantly spoken in the state of Karnataka, India, with over 43 million native speakers according to the 2011 Census. It holds official language status in Karnataka and is one of the 22 scheduled languages of India. Kannada has a rich literary tradition dating back to the 9th century and uses the Kannada script, which evolved from the Kadamba script and bears resemblance to Telugu. The language features agglutinative morphology. Nouns are marked for gender, number, and case, while verbs are inflected for tense, aspect, mood, and agreement with the subject. Kannada shares linguistic features with other Dravidian languages but maintains unique phonological, syntactic, and lexical characteristics.

\subsubsection*{Telugu}
Telugu is a major Dravidian language spoken predominantly in the Indian states of Andhra Pradesh and Telangana, with over 81 million native speakers as per the 2011 Census, making it the fourth most spoken language in India. It is one of the 22 scheduled languages of India and has official language status in both states. Telugu boasts a literary history dating back to at least the 11th century, with inscriptions and classical poetry. The language is written in the Telugu script, which evolved from the Brahmi script and is closely related to the Kannada script. Telugu exhibits agglutinative morphology and has rich verbal inflection for tense, aspect, mood, number, and person, and nouns are marked for number and case. Though it shares core grammatical features with other Dravidian languages, Telugu has distinct phonological and lexical traits influenced by Sanskrit and Prakrit over centuries.



\subsubsection*{Punjabi}

Punjabi is an Indo-Aryan language primarily spoken in the Indian state of Punjab and in Pakistan’s Punjab province. It is one of the few Indo-Aryan languages that is tonal, a feature more typical of some Tibeto-Burman languages. According to the 2011 Census of India, Punjabi has over 33 million speakers in India, and globally it is spoken by more than 125 million people, making it one of the most widely spoken languages in the world. In India, Punjabi is written using the Gurmukhi script, while in Pakistan it is typically written in Shahmukhi, a variant of the Persian script. The language exhibits typical Indo-Aryan features such as postpositions, gendered nouns, and a rich system of inflectional morphology. Its grammar includes two numbers (singular and plural), two genders (masculine and feminine), and multiple cases. Punjabi verbs are conjugated for tense, aspect, mood, person, number, and gender. Tonal distinctions are phonemic, usually marked by pitch and aspiration in speech, though not represented in the script. Punjabi shares a significant lexical and syntactic overlap with Hindi and Urdu but maintains distinct phonological and cultural characteristics, including a vibrant oral and literary tradition rooted in Sufi and Sikh heritage.

\subsubsection*{Odia}

Odia is an Indo-Aryan language spoken primarily in the Indian state of Odisha, where it serves as the official language. With over 40 million speakers according to the 2011 Census of India, Odia holds the status of one of the classical languages of India, recognized for its rich literary heritage. The language is written in the Odia script, which is an abugida, sharing similarities with other Brahmic scripts but with distinct features such as rounded shapes and unique vowel notation. Odia's grammatical structure includes three genders (masculine, feminine, and neuter), two numbers (singular and plural), and three cases (nominative, accusative, and instrumental). The verb system in Odia is highly inflected, marking distinctions for tense, aspect, mood, and person. One of the notable features of Odia is its phonological system, which includes 12 vowels and 34 consonants, with a rich system of vowel harmony and nasalization. While it shares common lexical features with other Indo-Aryan languages like Bengali and Assamese, Odia has maintained a unique phonological and morphological identity. The language has a deep cultural significance, reflected in its long tradition of classical music, dance, and literature, including works of great poets such as Sarala Das and Fakir Mohan Senapati.

\subsubsection*{Urdu}

Urdu is an Indo-Aryan language spoken primarily in Pakistan and parts of India, serving as one of the official languages of both nations. It is the native language of approximately 70 million people, with many more speakers using it as a second language. Urdu evolved during the Mughal era, absorbing vocabulary from Persian, Arabic, and Turkish, which gives it a distinct literary style and linguistic heritage compared to its close relatives like Hindi. Urdu is written in the Persian script, a variant of the Arabic script, and is known for its calligraphic beauty. The verb system is highly inflected, with distinctions in tense, aspect, and mood. Urdu's phonological system is rich, with a wide variety of consonants and vowels, many of which are borrowed from Persian and Arabic. The language shares a significant portion of its vocabulary and grammar with Hindi, though its script, literary traditions, and vocabulary set it apart. Urdu literature, particularly poetry, holds immense cultural value, with poets like Mirza Ghalib and Allama Iqbal shaping the language's literary prestige. The language serves as a bridge for communication across diverse communities, fostering cross-cultural dialogue in South Asia.

\subsubsection*{Kashmiri}

Kashmiri is an Indo-Aryan language primarily spoken in the Kashmir Valley in India and parts of Pakistan. It is the mother tongue of approximately 7 million people, though it faces challenges in terms of preservation due to the dominance of Hindi and Urdu in the region. Kashmiri belongs to the Dardic group of the Indo-Aryan languages and is notably distinct from its neighboring languages like Punjabi and Dogri. The language has a rich history, influenced by Sanskrit, Persian, Arabic, and Tibetan due to centuries of trade and cultural exchange in the region. Kashmiri is written in both the Perso-Arabic script and the Devanagari script, with the former being more commonly used in literary contexts. The phonology of Kashmiri is unique, with a system of retroflex consonants, and it includes a series of guttural sounds and tonal distinctions. The language employs an ergative-absolutive syntactic structure, which is rare among Indo-Aryan languages, making it fascinating from a linguistic perspective. Kashmiri has three grammatical genders (masculine, feminine, and neuter), and its verb system marks distinctions of tense, aspect, and modality. The language is rich in literary traditions, particularly poetry, with famous poets like Noor-ud-Din Noorani contributing to its cultural legacy.

\subsubsection*{Hindi}

Hindi is an Indo-Aryan language and one of the most widely spoken languages in India and across the world. It is part of the larger Indo-European language family and is the official language of India, alongside English. Hindi has around 44\% of India’s population as speakers, amounting to more than 500 million people. It is written in the Devanagari script, and its vocabulary draws heavily from Sanskrit, though it also incorporates elements from Persian, Arabic, and Turkic due to historical influences. Hindi has a complex phonetic system that includes both aspirated and unaspirated consonants, and its grammar is characterized by its use of postpositions, gendered nouns, and verb conjugations. The language features a rich literary tradition, with classical works dating back to medieval poetry and contemporary works in modern Hindi literature. Hindi also serves as a lingua franca for millions of speakers across the Indian subcontinent, facilitating communication between speakers of various regional languages.

\subsubsection*{English}

English is a Germanic language that originated in England and is now a global lingua franca, spoken by millions as a first or second language. It is the dominant or official language in many countries, including the United States, the United Kingdom, Canada, Australia, and New Zealand, and it serves as a primary medium for global communication in business, science, technology, and diplomacy. English belongs to the Indo-European language family and has undergone significant evolution, from Old English, influenced by Old Norse, to Middle English, shaped by French during the Norman Conquest, and finally to Modern English. Its vocabulary is vast, having borrowed words from a variety of languages, including Latin, French, and German. English is known for its flexible syntax, relatively simple conjugation system, and rich array of tenses. Phonologically, English is notorious for its lack of consistency between spelling and pronunciation, with many exceptions and irregularities. Despite being spoken globally, English has many dialects and accents, with notable variations in vocabulary, pronunciation, and grammar across regions. It is the primary language of international business and diplomacy, and its influence continues to grow in fields such as technology, entertainment, and academia, making it one of the most important languages in the world today.

\subsection{Domains} 

The collected parallel corpora spans over a number of domains of significant importance. The distribution of the collected corpora over the different domains is shown in table \ref{tab:splits}. 

\subsubsection*{Governance}

The government domain includes official records, policies, and public service announcements. Language technology in this domain supports administrative efficiency, citizen engagement, and policy transparency. Processing such data ensures multilingual communication, promoting inclusive governance and access to essential public services.

\subsubsection*{Healthcare}

The healthcare domain is vital for developing NLP tools that improve medical communication, patient care, and health information accessibility. Language models trained on healthcare data can assist in diagnosis, telemedicine, and translating critical medical documents, ensuring inclusive access across linguistic and regional barriers.

\subsubsection*{General}

The general domain comprises everyday communication, news, and informal discourse. It reflects natural language use and social context, offering diverse linguistic patterns for training robust models. Incorporating this domain enhances a system’s adaptability to real-world applications and user interactions.

\section{Related Works}

The development of parallel corpora, a critical resource for machine translation, is a challenging task undertaken through several methods. The most conventional is the manual approach, where source text is meticulously translated by human experts. This method remains the gold standard for quality, as it effectively captures linguistic nuance, cultural context, and complex idiomatic expressions that automated methods often miss. While it produces the highest fidelity data, it is notoriously expensive and time-consuming. Major initiatives like the Linguistic Data Consortium (LDC) serve as primary distributors for such corpora, including the Arabic Broadcast News Parallel Text. Similarly, the Indian Language Corpora Initiative (ILCI) was a significant undertaking where sentences were collected and manually translated for various Indian languages from official promotional materials to foster digital content creation \cite{jha-2010-tdil}. Many modern corpora also rely on meticulous human curation from diverse, high-quality sources like the IIT-Bombay Hindi-English Parallel Corpus (IITB-hi-en) \cite{kunchukuttan-etal-2018-iit}, OPUS \cite{tiedemann-2012-parallel}, HindEn \cite{bojar-etal-2014-hindencorp}, and TED talks \cite{abdelali-etal-2014-amara}. A concrete example of such painstaking manual effort is the collection of Sindhi data for this work, where text was manually keyed in from textbooks and extracted from legacy Pagemaker files—a process involving significant technical challenges in converting outdated fonts and formats before the data could be carefully filtered. However, not all human-contributed data is equal; it requires rigorous vetting. The WAT-ILMPC corpus \cite{nakazawa-etal-2017-overview}, for instance, derived from user-contributed subtitles, was found to be noisy and code-mixed \cite{philip-etal-2019-cvits}, a common issue with crowdsourced content that highlights the absolute necessity of rigorous curation.

To accelerate corpus creation and overcome the scalability limitations of \textbf{manual translation}, several \textbf{automated} and \textbf{semi-automated} methods are employed. The \textbf{sentence alignment approach} is a foundational technique used to automatically align sentences with their translations in existing parallel documents, such as a book and its published translation. Early statistical algorithms relied on simple heuristics, such as the principle that shorter sentences in a source text likely correspond to shorter sentences in the target text, using sentence length in words or characters as a proxy for alignment \cite{gale-church-1993-program}. Later methods improved accuracy by incorporating lexical information, using bilingual dictionaries to anchor the alignment process \cite{chen-1993-aligning}. This technique was effectively scaled using modern neural methods to create a 407K sentence-pair corpus for 10 Indian languages by aligning crawled web data from official sources like the Press Information Bureau \cite{siripragada2020multilingual}. Complementing this is the \textbf{web mining approach}, which involves discovering and scraping naturally occurring parallel content from multilingual websites. This method was instrumental in building IndicCorp, a large monolingual corpus of nearly 9 billion tokens scraped from news, magazines, and books \cite{kakwani2020indicnlpsuite}. A third powerful technique is the \textbf{machine translation approach}, commonly known as Machine Translation Post-Editing (MTPE). In this workflow, a monolingual text is first translated by an MT system to produce a draft, which is then refined by human experts. This significantly reduces the cognitive load on translators and saves considerable time. The ML4HMT corpus, for example, was built on this principle and comprises annotated MT outputs for English-to-German, Spanish, and Czech \cite{avramidis-etal-2012-richly}.

A distinct and increasingly vital method for expanding corpora, especially for low-resource pairs, is through \textbf{synthetic data generation}. This involves creating new parallel sentence pairs programmatically rather than through direct translation. While techniques like back-translation are common, the most prominent strategy for multilingual settings is using a pivot language. The logic of this approach is to bridge two languages via a third, high-resource language, typically English. The most significant example of this technique in the Indic context is the Samanantar corpus \cite{ramesh2022samanantar}. The project first compiled a large corpus of 49.7 million sentence pairs between English and 11 Indian languages. This resource was then used to synthetically generate an enormous new corpus of 83.4 million sentence pairs for 55 direct Indic-to-Indic language pairs by using English as an intermediary. This pivoting strategy dramatically increases the volume of parallel data, but it is not without risks, such as the potential for error propagation from the initial translation steps. Despite this, the resulting large-scale datasets have been instrumental in training powerful multilingual models like IndicTrans, which was released alongside the corpus to facilitate direct translation between Indian languages without relying on English, marking a significant step towards building a more inclusive digital ecosystem.


\section{IL-IL Corpora Development}
\subsection{Data Sources}

When using data sources for creating a parallel corpus, it's very important to consider copyright laws and intellectual property rights. Many texts which are published by governments or international organizations were either in the public domain or available under open licenses are suitable for use. However, there are websites, commercial publications and literary works which are often protected by copyright. Unauthorized use of these materials can lead to legal issues and ethical concerns. To avoid potential infringement, we sought for permission from the copyright owners. Also we used content available under licenses that allow for such use, like Creative Commons licenses. Proper attribution and adherence to the terms of these licenses are also crucial. As mentioned earlier, we targeted selecting monolingual text data mainly from two domains, i.e Governance and Health domain. We handpicked and ensured that selected monolingual data should capture all language nuances, diversity, dialects and vocabulary to capture the full spectrum of domain data.  Details of each sources is duly attributed and can be found in Annexure A.

    

    

\subsection{Corpora Cleaning/Filtering}

Since most of the collected corpora are web-scraped, preprocessing is essential before they are ready for translation or annotation. The preprocessing steps outlined below are applicable to both monolingual and parallel corpora. First, it is crucial to remove special or unwanted characters, multiple symbols, and repeated punctuation marks that often occur in scraped data. For example, text fragments like ``Hello!!!'', ``he said ---and went.'', or concatenated words such as { \dn EvkSpo{\qva}ko\7{c}nnA }
need to be cleaned for consistency and clarity. Sentences containing invalid characters or characters outside the valid Unicode range for a given language are identified and removed. Additionally, duplicate sentences or multi-word sequences that are not appropriately connected (e.g., lacking hyphens) are eliminated to reduce redundancy. Another important step involves handling sentences that contain extremely long words (e.g., ``hascongratulatedtheIndianCricket'') or exhibit poor word segmentation. Sentences that are unusually short (fewer than three words) or excessively long (over eighty words) are also filtered out to maintain a balanced dataset. Text normalization plays a key role in resolving inconsistencies and standardizing content. This includes harmonizing different forms of abbreviations such as ``S. No.'', ``Sr. No.'', ``Sl. No.'', and ``Sn. No.'' into a standard form like ``Serial No.'' Similarly, gender representations are converted from ``male'' and ``female'' to standardized tokens like ``M'' and ``F''. To improve the quality of the corpus, bulleted points, whether from ordered or unordered lists are removed. Accented characters are standardized by converting them into ASCII representations. Examples include transforming \textit{résumé}, \textit{café}, \textit{prótest}, \textit{divorcé}, \textit{coördinate}, \textit{exposé}, and \textit{latté} into their ASCII equivalents. Contractions are also expanded to their full forms to enhance standardization and avoid ambiguity. For instance, ``don’t'' is expanded to ``do not'', ``I’d'' becomes ``I would'', and ``you’re'' is rewritten as ``you are''. Furthermore, words found within brackets are extracted where necessary, and symbols with different encodings are normalized across the corpus. This includes multiple types of dash symbols, variations in double quotes, inconsistencies in Hindi numerical forms, and inconsistently encoded words. Overall, these preprocessing steps help in producing clean, uniform, and high-quality datasets that are suitable for downstream tasks like machine translation and linguistic annotation.

\subsection{Guidelines for Parallel Corpora}
Preparing clear and detailed guidelines for parallel corpora creation is essential to ensure consistency, quality, and reliability of the data. Guidelines help define the scope of data collection, text selection criteria, translation standards, annotation conventions, and formatting rules. They also minimize subjectivity among contributors, reduce errors, and make the corpus reproducible for future research. For this project, a detailed guideline document was prepared and circulated among the consortium members as a standard reference, ensuring that the parallel corpus is accurate, uniform, and useful for both academic and practical applications. Excerpts of that is given in Appendix~\ref{app:parallelcorpora}

\subsubsection{Involved Language Experts (Selection Process)}


    

    

    

    

\color{black}

The language experts were selected through a rigorous screening process. All relevant experts hold at least a Master’s degree in their respective languages, with many possessing prior experience in MeitY-funded projects and several holding Ph.D. degrees. After shortlisting qualified candidates, an online assessment was conducted to evaluate their proficiency in language-specific translation. Measures were taken to prevent any misuse of online translation tools such as Google Translate. Following the assessment, shortlisted individuals were interviewed and further evaluated by in-house linguists and language specialists to ensure their suitability for the annotation task.

    

    

    

    

\subsubsection{Post Editing tool}
The Post-Edit (PE) tool served an important component in our translation task workflows, particularly when working with machine translation (MT) output for large-scale multilingual corpora. It served as an interface for refining raw MT output to ensure that the final text is both accurate in meaning and fluent in the target language. The tool supported a collaborative model involving our stakeholders such as external agencies, freelancers, and in-house language experts.

\begin{figure*}[htbp]
    \centering
    \includegraphics[width=1\linewidth]{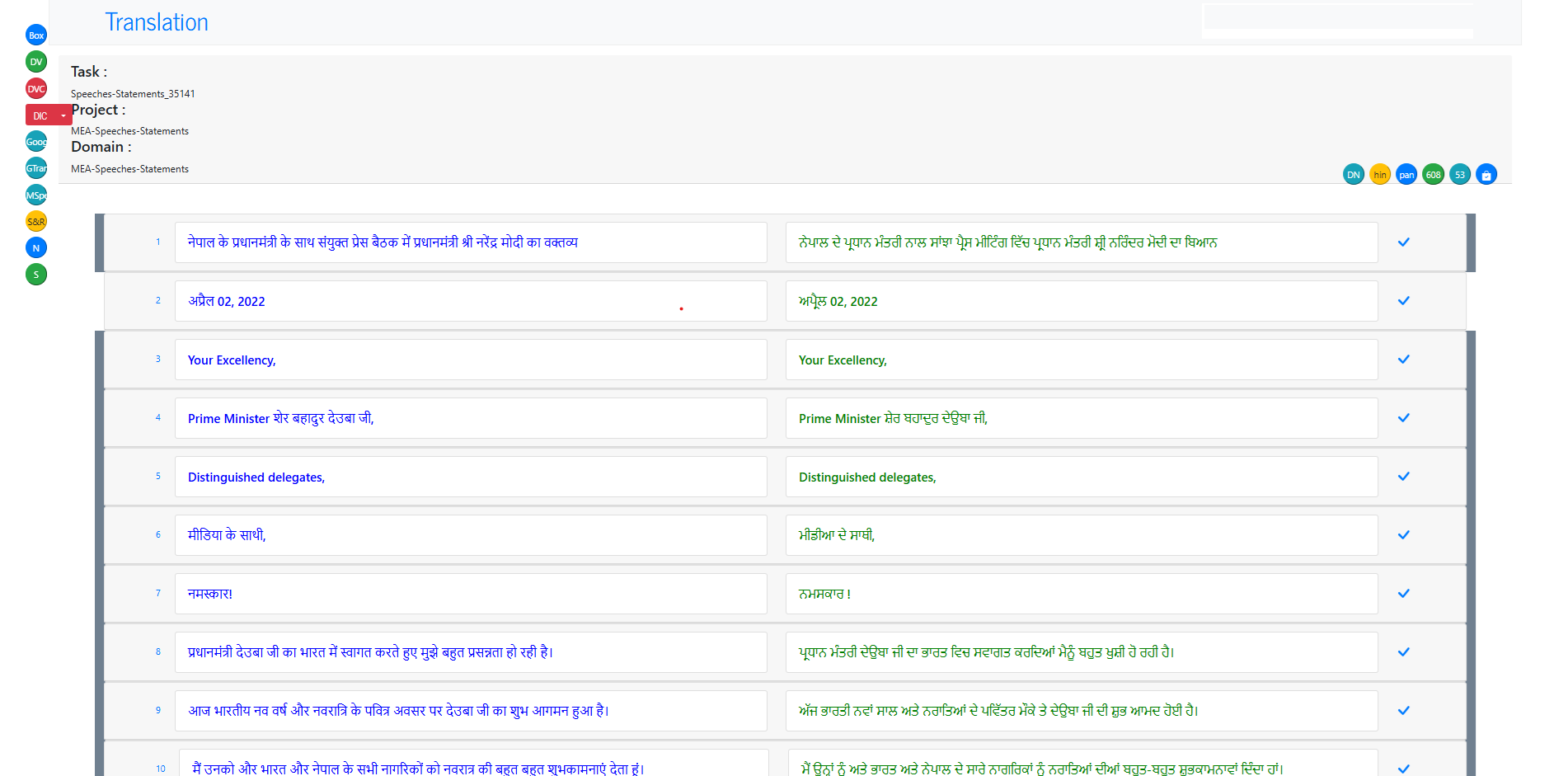}
    \caption{A snippet from the Post-Edit-Me platform}
    \label{fig:posteditme}    
\end{figure*}

In our setup, external agencies and freelancers were engaged to translate source material using either MT engines or manual methods. These initial translations were then brought into the post-editing pipeline, where in-house language vetters validated and refined the output. This two-tiered structure ensures both scalability and quality control — leveraging external resources for volume while maintaining linguistic consistency and accuracy through internal oversight. Figure \ref{fig:posteditme} shows PosteditMe tool developed under this project.

\section{Quality Check \& Comparison}
Ensuring the quality of a large corpus is crucial for developing machine translation models based on it. Two major types of approaches are followed for manual creation of corpora. The first one involves crowdsourcing which involves a large number of freelancers to develop the parallel corpora. Parallel corpora created using this approach often introduces lots of errors due to the variability of the workforce. The second kind of approach involves data creation using experts of particular language pairs. We adhere to the second approach. After the corpora has been developed, we employ two kinds of quality checks on it.
\subsection{Automatic}
In the process of developing a parallel corpora for Indian languages, ensuring the quality and accuracy of the data is paramount. To address this, we implemented an automatic sanity checker to identify and rectify common annotation errors observed during initial manual checks. The primary goal of this post-corpora creation effort was to enhance the overall quality of the dataset. Below, we detail the various types of errors handled by our sanity checker:

\begin{itemize}
    \item \textbf{New Line Detection}: Instances where either the source or target side of a language pair contains a new line character are flagged. This helps in maintaining the integrity of sentence alignment across the dataset.
    \item \textbf{Symbol-Only Sentences}: Sentences that contain only symbols and no words are identified. Such entries are generally considered noise and are removed to ensure meaningful translations.
    \item \textbf{Missing Sentences}: Cases where either the source or target sentence is missing from a language pair are detected. This ensures that each entry in the parallel corpora is complete and usable.
    \item \textbf{URL-Only Sentences}: Sentences that consist solely of URLs are flagged. Since URLs do not contribute to linguistic translation, these entries are removed to maintain the dataset's relevance.
    \item \textbf{Language Mismatch Detection}: Using the FastText model implemented via the FastLangID \footnote{\url{https://pypi.org/project/fastlangid/}} library, sentences that belong to languages other than the designated source and target languages are identified. While this tool cannot detect Dogri and Kashmiri, and can only indicate probable presence for other Indian languages, flagged sentences undergo manual evaluation to confirm their correctness. 
    \item \textbf{Identical Source and Target Sentences}: Instances where the source text is identical to the target text are flagged, as this indicates a lack of translation. Such entries are corrected or removed to ensure valid translation pairs.
    \item \textbf{Probable Wrong Alignment}: Cases where there is an apparent misalignment, such as an English-to-Hindi pair in a Hindi-to-English file, are detected. This is again managed using the FastLangID library, with the same limitations as previously mentioned. Flagged entries are manually reviewed to confirm and correct any mis-alignments.
\end{itemize}

By incorporating these automated checks, we significantly improved the quality of the parallel corpora, ensuring that it is a reliable resource for machine translation tasks. This approach not only streamlines the data validation process but also provides a robust framework for maintaining high standards in linguistic data collection and annotation.
\subsection{Manual}
After filtering out the noise from the developed parallel corpora through automatic sanity checks, we conduct quality check on the corpora. For performing this, we utilize the COMET \cite{rei-etal-2020-comet} model without the reference translation for each of the parallel sentences in the developed corpora. We utilize the reference less direct assessment score model for this task which internally uses XLM Roberta \cite{conneau2019cross} embeddings to represent each sentence. We categorize the data from each domain into 3 buckets after we obtain scores for all the parallel sentences:
\begin{enumerate}
    \item Group 1: Where the COMET scores are less than 0.6
    \item Group 2: Where the COMET scores are in the range of 0.6 to 0.8 (0.8 score is excluded)
    \item Group 3: Where COMET scores are higher than or equal to 0.8
\end{enumerate}
1000 sentences from each file are then manually validated from first group. Similarly 5 percent sentences from each file from second group are also manually verified. We assume that all the sentences in the third group are qualitatively better. For newer languages such as Sindhi, Kashmiri, and Dogri, the number of parallel sentences falling in the first group is comparatively higher than other languages as the base XLM Roberta model inside the COMET model does not support these languages. For most of other pairs, a significant percentage of the parallel sentences (~85\%) falls in the third group which indicate the nature of the translation. Following this exercise, we ensure the consistency and quality of the released corpora.

\begin{table}[H]
\centering
\begin{tabular}{@{}l|ccc|ccc|ccc@{}}
\toprule
\multirow{2}{*}{\textbf{Lang\_Pair}} & \multicolumn{3}{c}{\textbf{gen}} & \multicolumn{3}{c}{\textbf{gov}} & \multicolumn{3}{c}{\textbf{hlt}} \\
\cmidrule(lr){2-4} \cmidrule(lr){5-7} \cmidrule(lr){8-10}
 & \textbf{train} & \textbf{dev} & \textbf{test} & \textbf{train} & \textbf{dev} & \textbf{test} & \textbf{train} & \textbf{dev} & \textbf{test} \\
\midrule
dg\_hi & 12411 & 500 & 500 & 6947 & 500 & 500 & & & \\
en\_hi & & & & 38790 & 500 & 500 & 10043 & 500 & 500 \\
en\_te & & & & 9976 & 500 & 500 & 17237 & 500 & 500 \\
gu\_hi & & & & 18850 & 500 & 500 & & & \\
hi\_dg & & & & 30359 & 500 & 500 & 4343 & 500 & 500 \\
hi\_en & & & & 42964 & 500 & 500 & 12187 & 500 & 500 \\
hi\_gu & & & & 26335 & 500 & 500 & 4899 & 500 & 500 \\
hi\_kn & & & & 27531 & 500 & 500 & 16351 & 500 & 500 \\
hi\_ks & 21103 & 500 & 500 & & & & & & \\
hi\_or & & & & 24291 & 500 & 500 & 9387 & 500 & 500 \\
hi\_pa & & & & 30373 & 500 & 500 & 11328 & 500 & 500 \\
hi\_sd & 21548 & 500 & 500 & & & & 13233 & 500 & 500 \\
hi\_te & & & & 8061 & 500 & 500 & 11911 & 500 & 500 \\
hi\_ur & 8956 & 500 & 500 & 9929 & 500 & 500 & 5271 & 500 & 500 \\
kn\_hi & & & & 16040 & 500 & 500 & 19148 & 500 & 500 \\
ks\_ur & 2606 & 500 & 500 & & & & & & \\
or\_hi & & & & 11581 & 500 & 500 & 18308 & 500 & 500 \\
pa\_hi & & & & 22098 & 500 & 500 & 22532 & 500 & 500 \\
sd\_hi & 3499 & 500 & 500 & & & & & & \\
te\_en & & & & 5527 & 500 & 500 & 6008 & 500 & 500 \\
te\_hi & & & & 4405 & 500 & 500 & 18246 & 500 & 500 \\
ur\_hi & 27791 & 500 & 500 & 8938 & 500 & 500 & 6259 & 500 & 500 \\
ur\_ks & 22820 & 500 & 500 & & & & & & \\
\bottomrule
\end{tabular}
\caption{HimangY Parallel Corpora Stats}
\label{tab:splits}
\end{table}

\section{CorIL}
The parallel data are stored in files in standard text format where each source and target sentence is separated by a tab character. Most of the parallel corpora is composed of sentences from health and governance domains except the low resource languages such as Dogri, Kashmiri, and Sindhi. For these languages, the data splits are from general domain. The dev and test sets for each language pair consist of 500 sentences in each domain. We ensure that there is no overlapping of sentences across the splits in each language pair. For checking the overlapping, we follow two approaches: a. exact matching, and b. cosine similarity using COMET \cite{rei-etal-2020-comet} embeddings. The first approach removes all the duplicate parallel sentences. The second approach identifies semantically similar sentences. These methods are employed to eliminate any data bias of the trained MT models arising due to the presence of similar kinds of texts.  The details of all the splits for all language pairs are presented in Table~\ref{tab:splits}. A total of 713.5 K parallel sentences across 25 language pairs has been created as a part of this work.

\section{Machine Translation}

\begin{table*}[h]
    \centering
    \includegraphics[width=0.9\linewidth]{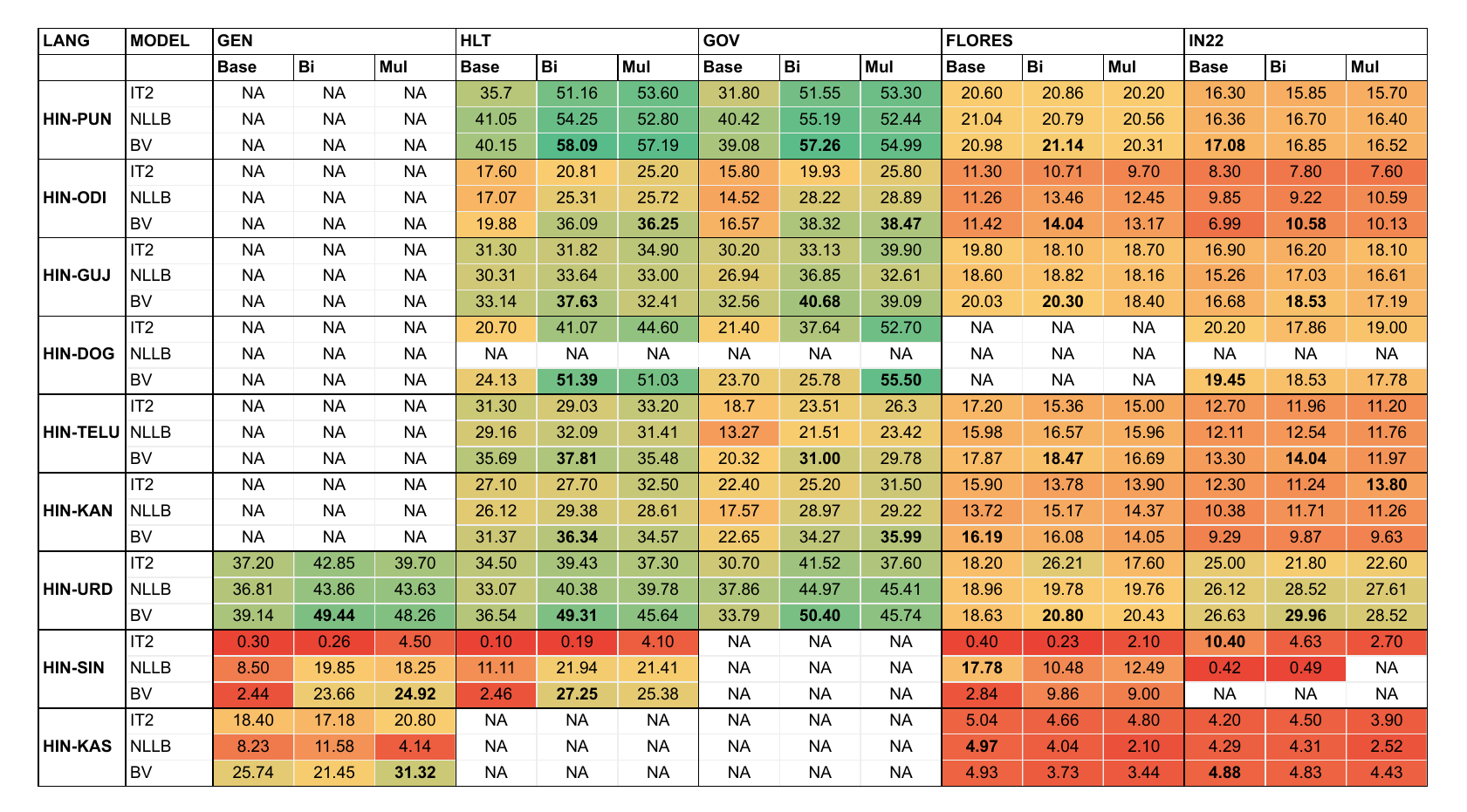}
    \caption{BLEU scores for machine translation between Hindi and Indian languages. The table compares the performance of Base, Bilingual (Bi), and Multilingual (Mul) fine-tuning for the IT2, NLLB, and BV models across five distinct domains (HLT, GOV, GEN, FLORES, and IN22).}
    \label{tab:hin_ind_bleu}
\end{table*}

In this section, we present the translation models that were fine-tuned using the CorIL corpus. We detail their underlying architectures and training paradigms, and evaluate their performance across various language pairs and domains. 

 \color{black}



 \begin{table*}[htb]
    \centering
    \includegraphics[width=0.9\linewidth]{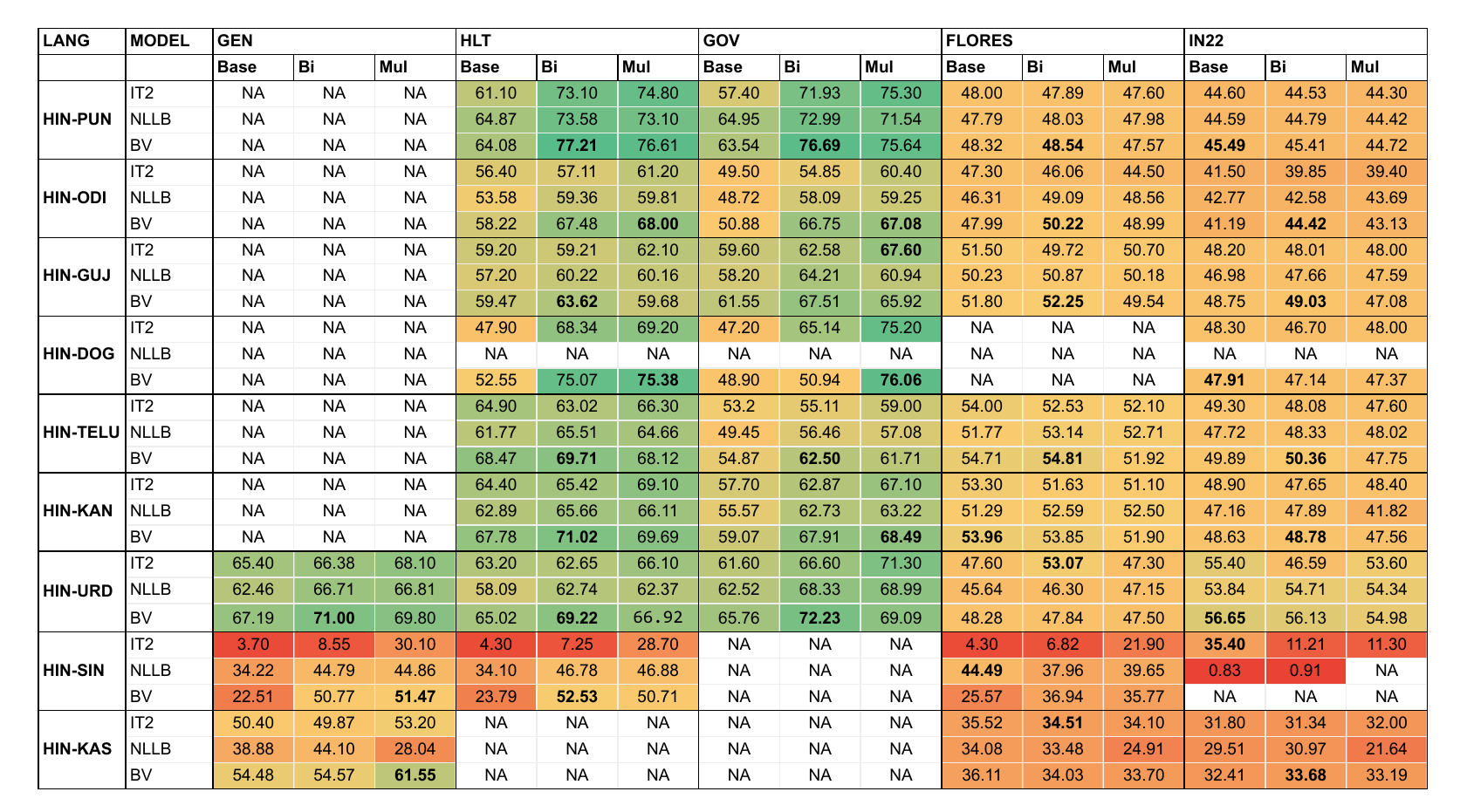}
    \caption{CHRF scores for machine translation between Hindi and Indian languages. The table compares the performance of Base, Bilingual (Bi), and Multilingual (Mul) fine-tuning for the IT2, NLLB, and BV models across five distinct domains (HLT, GOV, GEN, FLORES, and IN22).}
    \label{tab:hin_ind_chrf}
\end{table*}

\subsection{Model type}

We have conducted multiple experiments using our corpora with state-of-the-art machine translation models tailored for Indic languages, specifically No Language Left Behind (NLLB), IndicTrans2 and Bhasaverse \cite{mujadia2025bhashaversetranslationecosystem}. All these models have been trained on a vast array of languages and have demonstrated superior performance compared to other multilingual models for Indian languages. In addition to these, we have also trained our own translation model from scratch using the CorIL corpus combined with publicly available Indic parallel corpora.

No Language Left Behind (NLLB), developed by Meta AI, is a multilingual machine translation model designed to support over 200 languages, including numerous low-resource languages. The model employs a Sparsely Gated Mixture-of-Experts (MoE) architecture, enabling it to scale effectively while maintaining high translation quality. NLLB's training involved innovative data mining techniques to curate extensive datasets for low-resource languages, and it was evaluated using the FLORES-200 benchmark, achieving a 44\% BLEU improvement over previous state-of-the-art models . For practical deployment, Meta AI released distilled versions of NLLB, such as the NLLB-200-distilled-600M, which utilizes a dense BART-like architecture with sinusoidal positional embeddings and encoder-decoder attention mechanisms .

IndicTrans2, developed by AI4Bharat, is the first open-source transformer-based multilingual NMT model that supports high-quality translations across all 22 scheduled Indic languages. The model adopts script unification wherever feasible to leverage transfer learning through lexical sharing between languages. IndicTrans2's training leveraged the Bharat Parallel Corpus Collection (BPCC), the largest publicly available parallel corpora for Indic languages, comprising approximately 230 million bitext pairs. The model architecture is based on the transformer framework and includes variants with up to 1.12 billion parameters, such as the ai4bharat/indictrans2-en-indic-1B model.

BhashaVerse, developed by IIIT-Hyderabad’s Language Technologies Research Centre, is a multilingual, multi-task sequence-to-sequence (encoder-decoder) model covering 36 Indian subcontinent languages, including Assamese, Odia, Kashmiri (in multiple scripts), Tulu, Bodo, Hindi, Urdu, Bengali, Tamil, Sinhala, Santali, Khasi, and more. It is built on 10 billion parallel sentence pairs, spanning all 36 × 36 language directions, and integrates millions of additional data points for translation subtasks such as post-editing, grammar correction, error identification, and quality estimation. BhashaVerse employs multi-task learning, unifying tasks—machine translation, grammar correction, error detection, automatic post-editing, and quality estimation—within the same framework. It also supports discourse-level translation, enabling context-aware handling of paragraphs rather than just isolated sentences. Moreover, BhashaVerse incorporates both reference-based metrics (BLEU, CHRF3, COMET) and reference-free evaluation to assess translation quality. 


\color{black}
These models represent significant advancements in machine translation for Indic languages, offering robust solutions for translating between English and the diverse linguistic landscape of India. 

\begin{table*}[htb]
    \centering
    \includegraphics[width=0.9\linewidth]{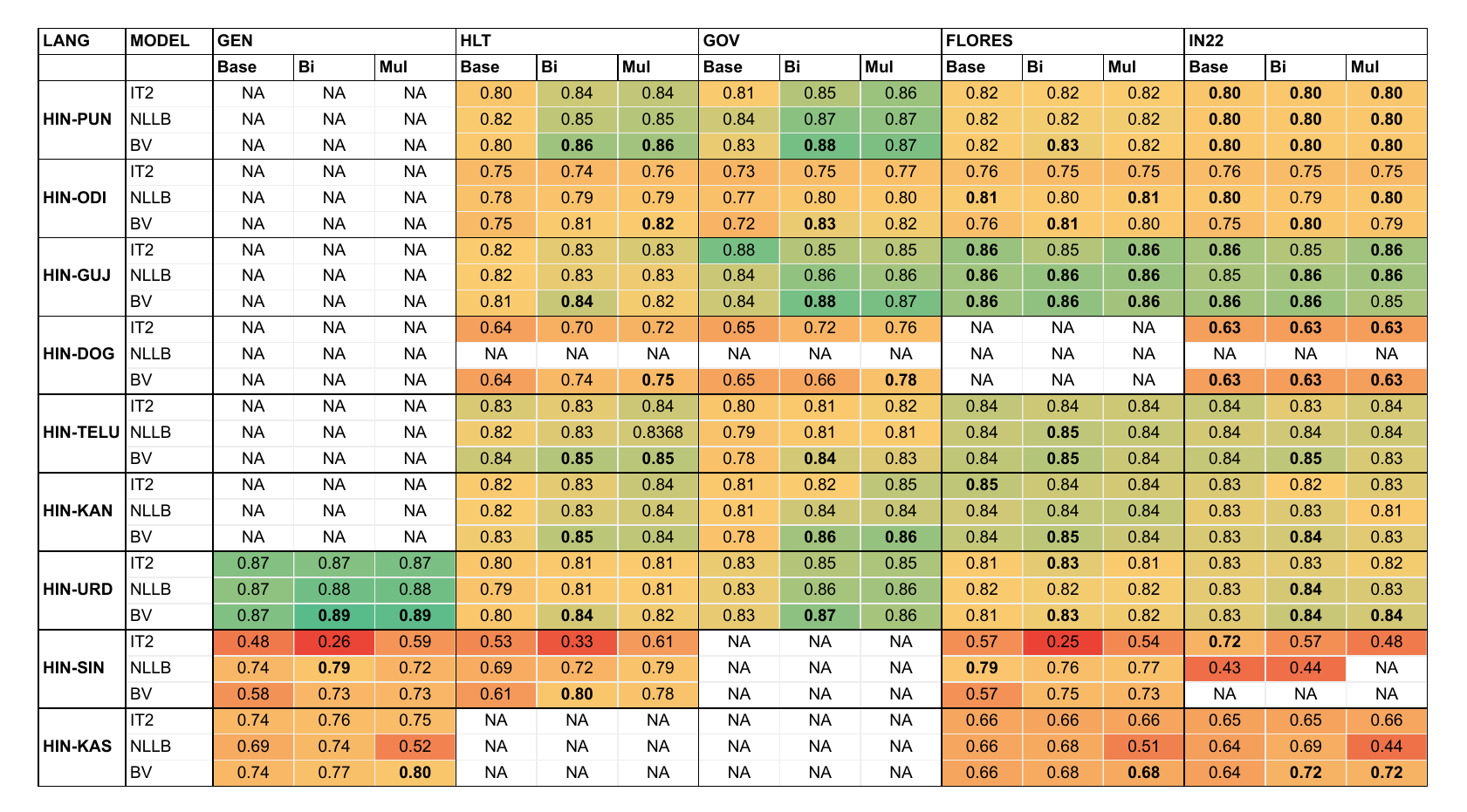}
    \caption{COMET scores for machine translation between Hindi and Indian languages. The table compares the performance of Base, Bilingual (Bi), and Multilingual (Mul) fine-tuning for the IT2, NLLB, and BV models across five distinct domains (HLT, GOV, GEN, FLORES, and IN22). It is important to note that Kashmiri, Dogri and Sindhi languages are not supported by COMET, hence their scores are highly unreliable.}
    \label{tab:hin_ind_comet}
\end{table*}

\subsection{Domains and Language Pairs}

 To adapt state-of-the-art machine translation models to the specific linguistic and domain requirements of the Himangy Project, we conducted extensive finetuning experiments using the project's curated datasets. The Himangy Project aims to develop bidirectional machine translation systems for English $\leftrightarrow$ Hindi, English $\leftrightarrow$ Telugu, and nine other Indian language pairs (Hindi $\leftrightarrow$ Punjabi, Telugu, Urdu, Gujarati, Kannada, Odia, Kashmiri, Sindhi, and Dogri).
 
 The dataset was partitioned into into training, development and test sets, ensuring a balanced distribution across language pairs and domains as can be seen in the table \ref{tab:splits}. The dataset encompasses three primary domains namely, general, government and healthcare. This stratification allows for targeted finetuning and evaluation of models within specific contextual settings. Our experiments covered a total of 23 unique translation directions, structured as follows: 
 \begin{enumerate}
    \item \textbf{Indic} $\leftrightarrow$ \textbf{Hindi}: Bidirectional translation experiments between Hindi and nine Indic languages: Punjabi, Odia, Gujarati, Telugu, Kannada, Dogri, Urdu, Sindhi, and Kashmiri.
    \item \textbf{English} $\leftrightarrow$ \textbf{Indic:} Bidirectional translation experiments between English and two Indic languages: Hindi and Telugu.
    \item \textbf{Hindi} $\rightarrow$ \textbf{Kashmiri:} A unidirectional translation experiment from Hindi to Kashmiri.
\end{enumerate}
 This comprehensive coverage ensures the evaluation of models across a diverse set of linguistic scenarios, including low-resource language pairs and varying script systems.

 \subsection{Finetuning strategies}

 We employed two primary finetuning strategies to assess the adaptability and performance of the base models (NLLB \cite{nllbteam2022languageleftbehindscaling}, IndicTrans2 \cite{gala2023indictrans2highqualityaccessiblemachine} and Bhashaverse\cite{mujadia2025bhashaversetranslationecosystem}). Models were finetuned on domain-specific subsets of the training data. For instance, the IndicTrans2 model was finetuned using the Health domain's Hindi-Punjabi training and development sets and evaluated on the corresponding test set. This process was replicated across all language pairs and domains using all the three models. Experiments were conducted in both bilingual and multilingual finetuning settings to assess the impact of training configurations on model performance.
 
 For benchmarking purposes, domain-specific datasets were amalgamated to create a comprehensive training set. Base models (NLLB, IndicTrans2 and BhashaVerse) were finetuned on this combined dataset and evaluated using benchmark datasets such as FLORES-200 \cite{nllbteam2022languageleftbehindscaling} and IN22 General Test \cite{gala2023indictrans2highqualityaccessiblemachine}. In cases where only general domain data was available for certain language pairs, the general domain-finetuned models were directly evaluated on the benchmark datasets.

\begin{table*}[h]
    \centering
    \includegraphics[width=0.9\linewidth]{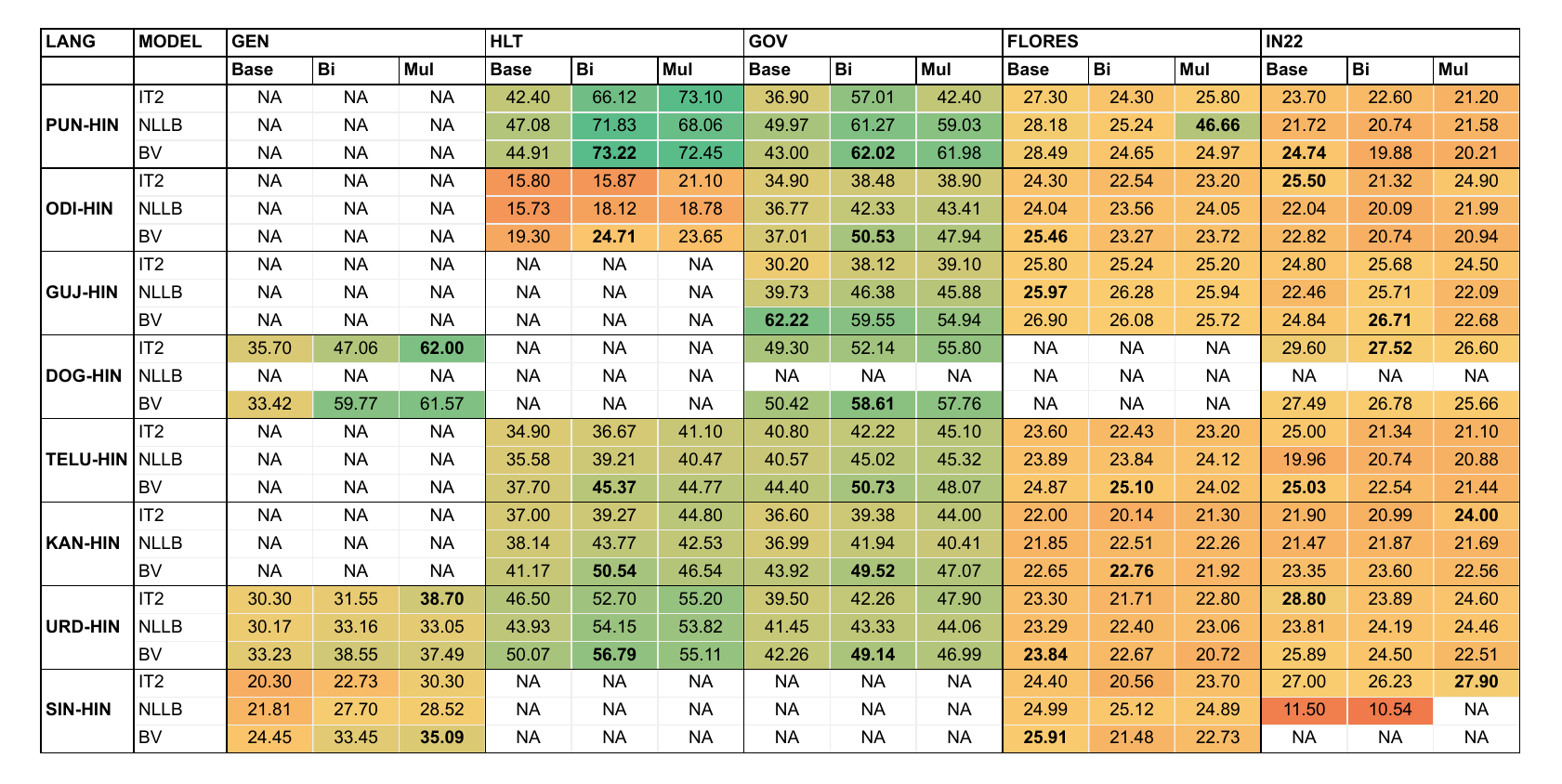}
    \caption{BLEU scores for machine translation between Indian languages and Hindi. The table compares the performance of Base, Bilingual (Bi), and Multilingual (Mul) fine-tuning for the IT2, NLLB, and BV models across five distinct domains (HLT, GOV, GEN, FLORES, and IN22).}
    \label{tab:ind_hin_bleu}
\end{table*}

 \subsection{Experimental Configurations}

 To systematically evaluate the performance of state-of-the-art machine translation models on the released dataset, we employed three prominent models: NLLB-200 \cite{nllbteam2022languageleftbehindscaling}, IndicTrans2 (IT2) \cite{gala2023indictrans2highqualityaccessiblemachine} and BhasaVerse\cite{mujadia2025bhashaversetranslationecosystem}. Each model was fine-tuned under both multilingual and bilingual settings to assess their adaptability across various linguistic scenarios. The configurations for each setup are detailed in the following subsections:

\subsubsection{NLLB multilingual/bilingual}
For both multilingual and bilingual configuration, we utilized the facebook/nllb-200-3.3B model, a 3.3 billion parameter multilingual machine translation model developed by Meta AI. To optimize for computational efficiency and resource constraints, we employed a combination of advanced quantization techniques and parameter-efficient fine-tuning strategies.

Specifically, we applied 4-bit NormalFloat (NF4) quantization with double quantization and bfloat16 computation. This approach, inspired by QLoRA \cite{dettmers2023qloraefficientfinetuningquantized}, significantly reduces memory usage while maintaining model performance. To enable efficient fine-tuning, we integrated Low-Rank Adaptation (LoRA) into the model. LoRA \cite{hu2021loralowrankadaptationlarge} introduces trainable rank decomposition matrices into each layer of the Transformer architecture, allowing for adaptation with a reduced number of trainable parameters.

In our configuration, we set the LoRA rank (r) to 8, the scaling factor ($\alpha$) to 16, and applied a dropout of 10\%. These adaptations were applied to the attention mechanisms, specifically the query, key, value, and output projections. We employed mixed precision training using fp16 to balance performance and resource utilization.

Training was conducted over 5 epochs with a batch size of 8, a learning rate of 2e-5, and a weight decay of 0.01. Model evaluation was performed at the end of each epoch to monitor performance and prevent overfitting. Translation quality was assessed using the BLEU score, CHRF and COMET. This configuration was applied consistently across all fine-tuning experiments, encompassing various language pairs and domains.

\subsubsection{IndicTrans2 multilingual/bilingual}

For the multilingual and bilingual fine-tuning experiments, we employed the IndicTrans2 model, specifically the distilled variant \textit{ai4bharat/indictrans2-indic-indic-dist-320M}. Due to computational constraints, we adopted this setup focussing on individual language pairs and allowing the model to specialize in the nuances of each pair. The training process utilized a batch size of 16 and spanned 3 epochs, with a maximum of 1,500 training steps. The learning rate, inititalized at 5e-4 followed an inverse square root scheduler, with a warmup phase of 150 steps. For parameter-efficient fine-tuning, Low-Rank Adaptation (LoRA) was integrated, targeting the query and key projection layers with a rank of 16, $\alpha$=32, and a dropout rate of 10\%. However, no quantization was employed since, we were already working with a distilled model to retain better performance. Evaluation was conducted every 100 steps. This configuration ensured that each bilingual model was finely tuned to capture the specific linguistic characteristics of the respective language pair, facilitating high-quality translations.

\begin{table*}[h]
    \centering
    \includegraphics[width=0.9\linewidth]{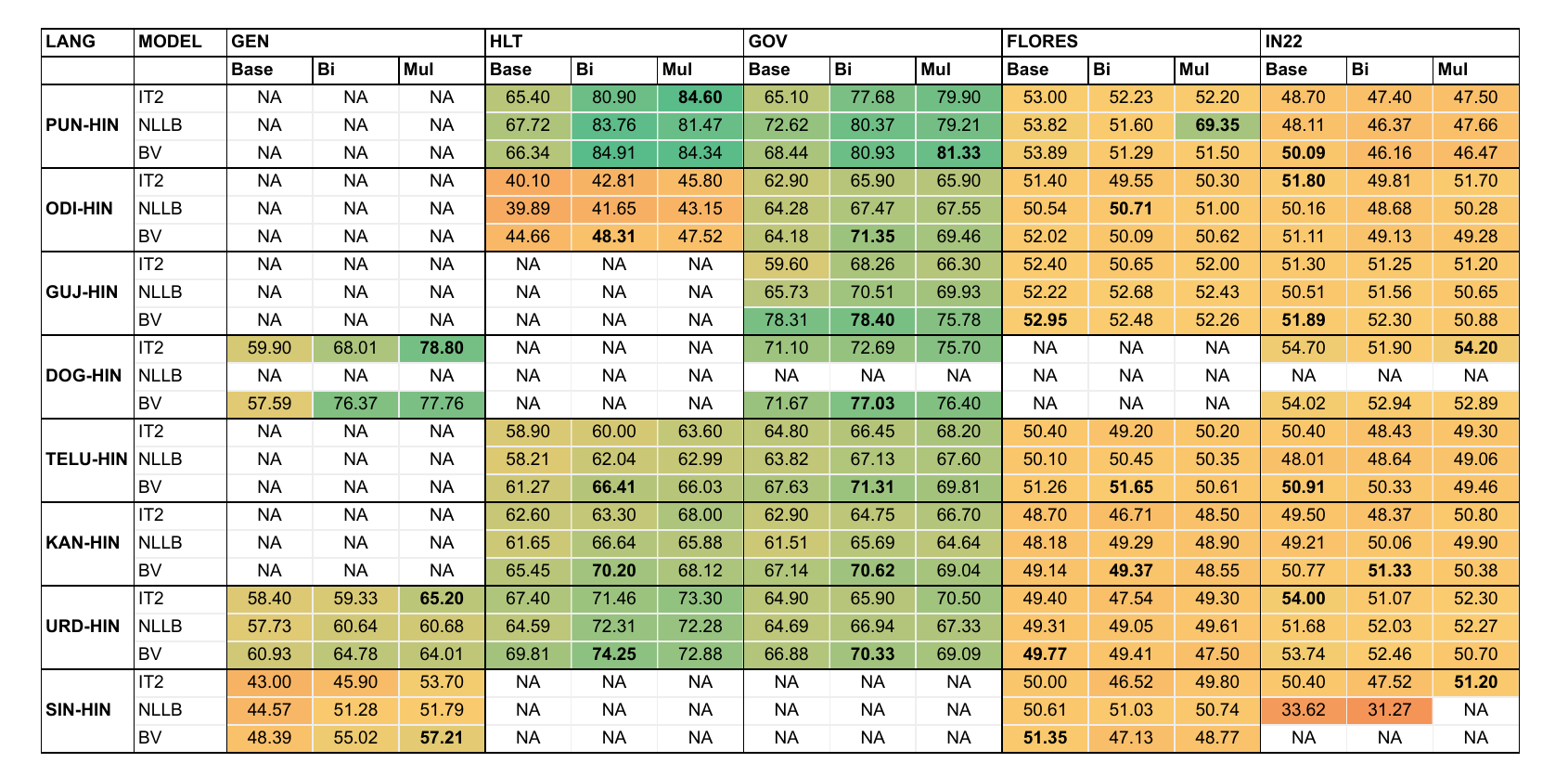}
    \caption{CHRF scores for machine translation between Indian languages and Hindi. The table compares the performance of Base, Bilingual (Bi), and Multilingual (Mul) fine-tuning for the IT2, NLLB, and BV models across five distinct domains (HLT, GOV, GEN, FLORES, and IN22).}
    \label{tab:ind_hin_chrf}
\end{table*}

\subsubsection{BhashaVerse multilingual/bilingual}

For the multilingual and bilingual fine-tuning experiments across 11 Indian language pairs, we employed the Bhashaverse model developed by LTRC, IIIT Hyderabad, and we have used the Fairseq framework with the custom transformer architecture. The last trained checkpoint was loaded and fine-tuned with a maximum of 101 epochs. The batch size is set to 4096 tokens; the optimization uses the Adam optimizer of (0.9, 0.98), with the smoothing factor of 0.1. The learning rate, starting with one at 5e-4, was optimized using an inverse square root scheduler and a warmup step of 4000 steps. Along with this, a dropout rate of 0.3 and a weight decay of 1e-4 were applied. Evaluation was conducted every epoch. Such a design made sure that every bilingual model had been refined lightly to reproduce the linguistic traits or peculiarities of the two specific languages in order to achieve translations of high quality. 

\begin{table*}[h]
    \centering
    \includegraphics[width=0.9\linewidth]{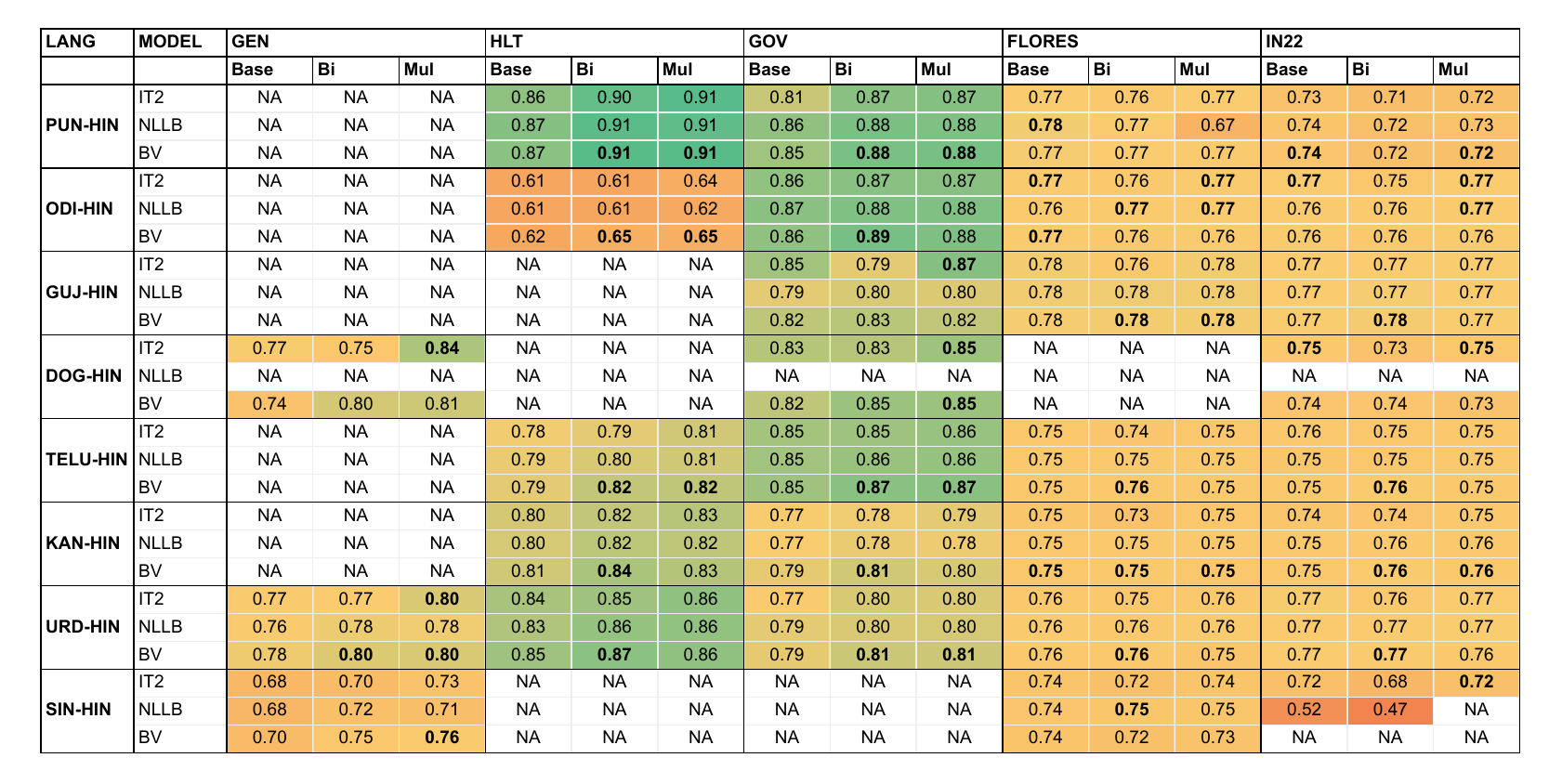}
    \caption{COMET scores for machine translation between Indian languages and Hindi. The table compares the performance of Base, Bilingual (Bi), and Multilingual (Mul) fine-tuning for the IT2, NLLB, and BV models across five distinct domains (HLT, GOV, GEN, FLORES, and IN22). It is important to note that Kashmiri, Dogri and Sindhi languages are not supported by COMET, hence their scores are highly unreliable.}
    \label{tab:ind_hin_comet}
\end{table*}

 \begin{table*}[htbp]
    \centering
    \includegraphics[width=0.9\linewidth]{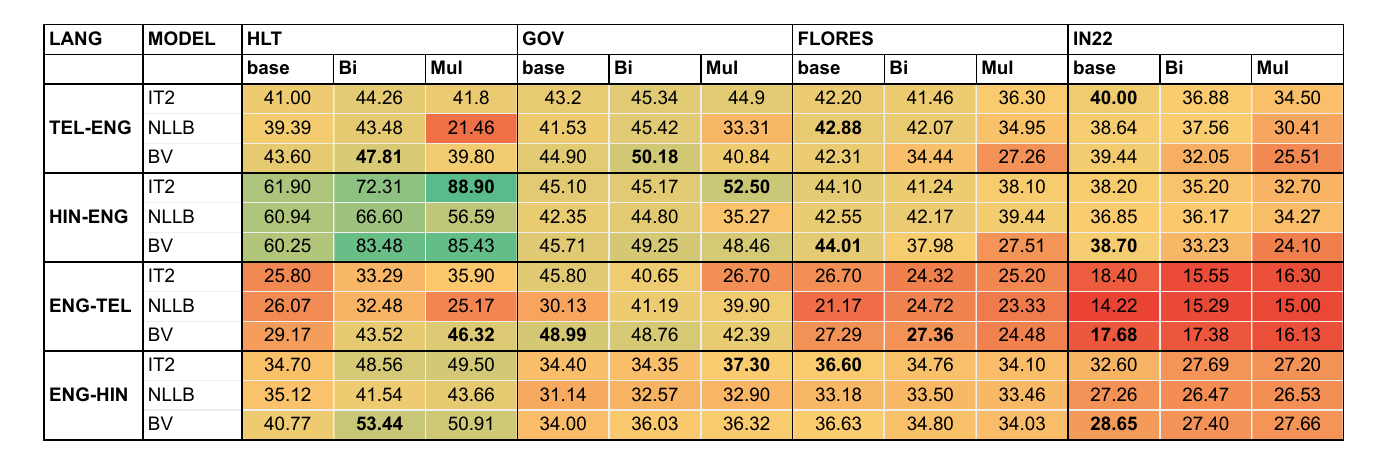}    
    \caption{BLEU scores for machine translation between English and Indian languages (Hindi and Telugu). The table compares the performance of Baseline, Bilingual (Bi), and Multilingual (Mul) fine-tuning for the IT2, NLLB, and BV models across four distinct domains (HLT, GOV, FLORES, and IN22).}
    \label{tab:eng_bleu}
\end{table*}

\color{black}

\section{Results and Discussions}

Experiments were conducted using the released corpora with both bilingual and multilingual finetuning strategies applied to three state-of-the-art translation models: IndicTrans2, NLLB and BhashaVerse. Finetuning consistently led to notable improvements over the baseline performance of all the three models on the domain-specific test sets derived from the released corpora. However, a contrasting trend was observed when the finetuned models were evaluated on public benchmark datasets, namely FLORES-200 devtest and IN22 gen. In these evaluations, the models exhibited a decline in performance compared to their baselines, highlighting the potential issue of domain overfitting. Similar trends were also noted for the models trained from scratch in the bilingual setting.


For language pairs with access to general-domain parallel data (e.g., Hindi-Urdu), additional finetuning was performed using the general-domain train set, followed by evaluation on FLORES-200 devtest and IN22 gen. These models showed marginal improvements on FLORES-200 devtest, indicating that exposure to general-domain data can partially mitigate the overfitting effects of domain-specific finetuning. Overall, the models achieved better performance in the governance domain compared to the healthcare domain. This may be due to the more standardized and familiar language used in governance-related texts, which allows better generalization. In contrast, healthcare datasets often contain highly specialized terminologies, posing greater challenges for finetuning and evaluation. Also bilingual fine-tuning generally proves to be the better performing and more reliable strategy across the majority of language pairs and domains.

The results are presented using a color-coded scheme: green indicates high values, yellow represents mid-range values, and red highlights lower values. This heatmap-style visualization facilitates quick and intuitive comparison of model performance across different domains and language pairs, making patterns and disparities easier to identify.

\begin{table*}[h]
    \centering
    \includegraphics[width=0.9\linewidth]{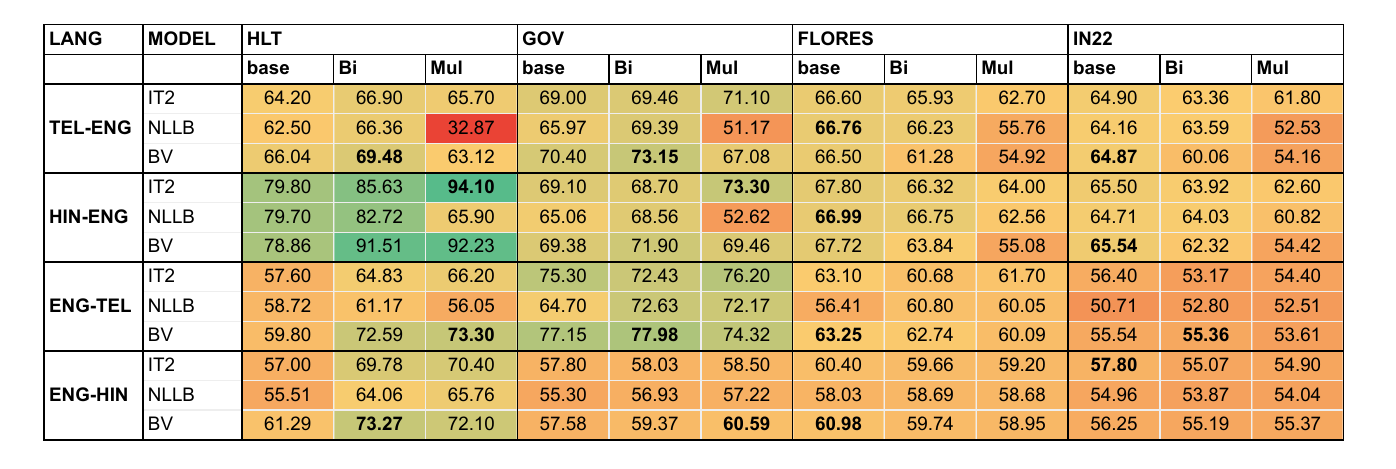}
    \caption{CHRF scores for machine translation between English and Indian languages (Hindi and Telugu). The table compares the performance of Baseline, Bilingual (Bi), and Multilingual (Mul) fine-tuning for the IT2, NLLB, and BV models across four distinct domains (HLT, GOV, FLORES, and IN22).}
    \label{tab:eng_chrf}
\end{table*}

\begin{table*}[h]
    \centering
    \includegraphics[width=0.9\linewidth]{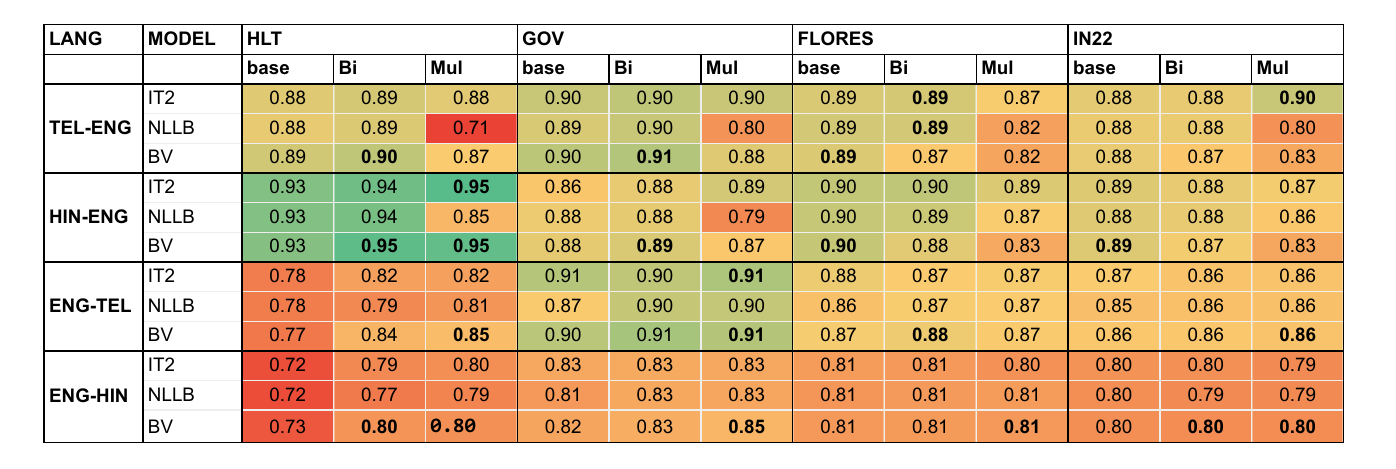}
    \caption{COMET scores for machine translation between English and Indian languages (Hindi and Telugu). The table compares the performance of Baseline, Bilingual (Bi), and Multilingual (Mul) fine-tuning for the IT2, NLLB, and BV models across four distinct domains (HLT, GOV, FLORES, and IN22).}
    \label{tab:eng_comet}
\end{table*}

\subsection{Domain wise Observations}

The primary motivation behind creating domain-wise data was to enable domain adaptation for the machine translation (MT) models. In the Hindi-to-Indic direction, fine-tuned models generally perform better in the government domain compared to the health domain. This can be observed through the relative prevalence of red-colored cells in the health domain, indicating lower performance. The reduced performance in the health domain can be attributed to its use of highly specialized vocabulary, in contrast to the more generic language seen in the government domain. Interestingly, performance in the general domain is also lower than that in the government domain—even when domain-specific fine-tuning data is available, as seen in the case of Sindhi and Kashmiri. This is counterintuitive, considering that the models are pre-trained on general-domain data. However, this observation must be contextualized by the fact that Sindhi and Kashmiri are low-resource languages and are likely underrepresented in the pre-training corpus. Conversely, Urdu, despite also undergoing general domain fine-tuning, performs better in the general domain compared to both health and government domains—likely due to its higher resource availability. Performance across the benchmarking dataset shows a noticeable decline, likely a result of domain-specific overfitting caused by fine-tuning on specialized data. Still, high-resource language pairs with general-domain availability, such as Urdu, show relatively better generalization across domains (see Table \ref{tab:hin_ind_bleu}).

In the Indic-to-Hindi direction, similar performance trends are observed. The government domain continues to be easier to translate than the health domain. The general domain continues to underperform, particularly for low-resource language pairs, where data scarcity affects quality. However, Urdu again stands out with relatively stronger performance. Benchmarking data performance is low across most language pairs, although in contrast to the Hindi-to-Indic direction, the models are more closely distributed in the Table \ref{tab:ind_hin_bleu}

For Hindi and Telugu pivoted around English, a different pattern emerges. In this setting, models fine-tuned on the health domain outperform those trained on the government domain. Nevertheless, performance on the benchmarking datasets remains comparatively low. Due to the abundance of training data in the English-Hindi direction, these language pairs exhibit the highest overall performance, as seen in Table \ref{tab:eng_bleu}.


\color{black}
\subsection{Language wise Observations}

Based on a detailed analysis, a clear performance hierarchy emerges among the models depending on the language's script. Although IndicTrans2 (IT2) is widely regarded as a state-of-the-art model for Indic languages, the provided data confirms that it struggles significantly with languages using the Perso-Arabic script, such as Urdu and Sindhi. For these languages, NLLB and BhashaVerse (BV) prove to be the better models. This is most evident with Urdu, where NLLB and BV consistently achieve positive scores, while IT2's performance is marked by highly negative results across all domains. Similarly, for Sindhi, NLLB's scores are considerably better than IT2's. This discrepancy could be attributed to IT2's training focus on Indic scripts, which limits its handling of Perso-Arabic scripts due to script divergence and under-representation in its training data. Conversely, NLLB's architecture as a massively multilingual model trained on a broader range of scripts likely enables better transfer learning, giving it a distinct edge. However, the data also shows that for Indic-script languages like Punjabi, IT2 performs exceptionally well, often matching or even exceeding NLLB's performance.

For both Hindi-Indic and Indic-Hindi language pairs, the best performance is observed with Punjabi, Gujarati, and Urdu, while Kashmiri and Sindhi consistently yield the lowest scores. These latter languages are relatively low-resource, even within the Indic language family, which likely contributes to their weaker performance. Additionally, both NLLB and IndicTrans2 lack Dogri in their pre-training datasets, and this absence is reflected in the notable drop in metric scores for Dogri, despite its affiliation with the Indo-Aryan language group. Overall, a general decline in performance is observed across all language pairs on the benchmarking dataset.

\subsection{Finetuning Specific Observations}


Bilingual ("Bi") fine-tuning is overwhelmingly the more effective and consistent strategy for improving machine translation performance. This approach consistently outperforms multilingual tuning across most language pairs, particularly in the Health (HLT) and Governance (GOV) domains. The impact of the fine-tuning strategy is least observable in the COMET scores, while the BLEU metric shows more prominent differences. The advantage is most dramatic in translations from Indian languages into English; for instance, in Telugu-to-English translation using the NLLB model, the BLEU bilingual score was 43.48, while the multilingual score was a dismal 21.46 as seen in Table \ref{tab:eng_bleu}. This indicates that a focused, two-language approach is generally a safer and higher-performing choice.

However, multilingual ("Mul") fine-tuning does show some benefits in specific, limited scenarios. It proves competitive, and sometimes superior, when translating from Hindi to other closely related Indo-Aryan languages like Punjabi and Odia. For example, the BhashaVerse model for HIN-ODI translation scored 36.25 with multilingual tuning versus 36.09 with bilingual tuning in the HLT domain. Similar trend was also observed for the other models and the GOV domain for HIN-ODI translations.

\section{Conclusion and Future Works}


\color{black}

In this paper, we introduced a high-quality parallel corpus for machine translation, encompassing English and 10 Indian languages across three critical domains: healthcare, government, and general. Notably, our dataset includes several low-resource and underrepresented languages such as Dogri, Kashmiri, and Sindhi, thereby addressing a significant gap in the existing resources available for Indian language translation.

To assess the utility and impact of this dataset, we conducted extensive experiments using both multilingual and bilingual fine-tuning strategies on leading translation models, including NLLB and IndicTrans2. Our experiments spanned both domain-specific and general-domain fine-tuning scenarios. Additionally, we benchmarked the performance of these fine-tuned models against their original baselines using widely accepted evaluation datasets such as FLORES-200 and the IN22 general test set. Our findings indicate an almost consistent improvement in translation performance when models are fine-tuned on our released corpus.

With the release of this dataset and our suite of empirical studies, we aim to provide a renewed impetus to the machine translation community. We believe this resource will facilitate more impactful and inclusive research, encouraging further exploration into effective modeling strategies for diverse Indian languages and their applications in real-world, domain-specific contexts.

\section{Acknowledgments}
The authors would like to attribute Shilalekh Publishers, Yojana Magazine, Publication Division, Central Health Education Bureau, National Health Portal, National Concil for Promotion of Sindhi Language, New Delhi, Mrs Shobha Lalchandani, Mrs. Tamana Lalwani, Mr. Deepak Lalwani (Ahmedabad), VikasPedia and Central as well State government websites for providing their consent to collect the monolingual data from their websites.
\bibliographystyle{ieeetr} 
\bibliography{bibliography.bib}
\clearpage
\appendix
\appendix
\section{About Parallel Corpora}
\label{app:parallelcorpora}
\addcontentsline{toc}{section}{Appendix A: About Parallel Corpora}

Parallel corpus is invaluable for Statistical Machine Translation (SMT) and Neural Machine Translation (NMT). Recently, NMT based systems have produced state-of-the-art (SOTA) results for many language pairs. These systems are trained on huge amount of parallel corpora of reasonable quality.

\subsection*{Application Areas of Parallel Corpora}
\begin{itemize}
    \item Statistical Machine Translation (SMT)
    \item Neural Machine Translation (NMT)
    \item Cross-Lingual Information Retrieval (CLIA)
    \item Speech Processing (SP)
\end{itemize}

\subsection*{Languages for Corpora Collection}
After reviewing the resources for all the languages, we categorize the languages into two types based on the availability of the required resources.

\paragraph{Category 1:}
For languages in category 1, parallel corpora and shallow parsing resources are available. The languages belonging to this category are:
\begin{itemize}
    \item Hindi
    \item Gujarati
    \item Punjabi
    \item Kannada
    \item Telugu
    \item Urdu
    \item Odia
\end{itemize}

\paragraph{Category 2:}
Languages of this category are extremely resource poor. These languages have little to very little digital presence. The languages belonging to this category are:
\begin{itemize}
    \item Dogri
    \item Kashmiri
    \item Sindhi
\end{itemize}

\subsection*{Corpora Collection}

\subsubsection*{a. Method of Data Collection and Creation}
For data collection, we will scrape the data from available websites for the required domains. The scraped data can either be monolingual or parallel. For this, popular Python packages like \texttt{BeautifulSoup}, \texttt{selenium}, etc. will be used.

For creating the parallel corpora, we will follow the three methods below:

\begin{enumerate}
    \item \textbf{End-to-End (E2E) Translation} -- Startups and other agencies will be engaged for creating translated corpora. In-house language editors will validate the translated data to ensure quality and consistency.
    \item \textbf{Back Translation} -- This is a well-known technique for improving NMT models by combining synthetic parallel data while training. In this approach a reverse NMT model trained on parallel data translates sentences from target-side monolingual data into the source language.
    \item \textbf{Collecting existing corpora and cleaning} -- Publicly available parallel corpora will be collected for different languages. State governments will also be contacted to provide other educational materials. Tools need to be developed to remove noise present in the existing parallel corpora which may severely impact the performance of the MT models.
\end{enumerate}

\subsubsection*{b. Nature of Corpora}
Categorization of corpora can either be based on domains or nature of language. The current project aims to create parallel corpora in two domains in the first two quarters. The details of the domains are given below:

\textbf{Governance:} Circulars, press releases, policy documents, government websites (Agriculture, External Affairs, Science and Technology, MeitY), annual reports.

\textbf{Healthcare:} Consent forms and information sheets; awareness material and pharma related documents.

\textbf{General Purpose.}

Based on the nature of language, the corpora can be divided into two categories:
\begin{itemize}
    \item Written
    \item Spoken
\end{itemize}

Linguistically, these texts are different. Spoken text contains speech disfluencies like repairs, repetitions, filler words, etc. In recent years, there has been a lot of emphasis on audio transcripts for language processing. Most of the current NMT systems are trained on written text. The performance of SOTA NMT systems usually degrades when evaluated on spoken text. To develop better generic models, we intend to train our models on spoken texts.

\paragraph{Sources for spoken text:}
\begin{itemize}
    \item Education video lectures
\end{itemize}

The sources for written text will be identified and some sources are mentioned above under the domains.

\subsubsection*{Representativeness of Corpus}
Average sentence length and type-token ratio measures are important indicators of linguistic properties of a corpus. Average sentence length (Number of tokens / Number of sentences) over a corpus indicates how representative the corpus is. The type-token ratio of text determines whether that text is of good quality or not. These indicators will be considered while selecting a corpus for parallel corpora.

\subsubsection*{c. Directory Structure \& Naming convention}
After collection of the corpora, it becomes highly essential to store them in a predefined directory structure. Proper naming convention for files is also required to make them uniquely identifiable. Index and store the corpus on the basis of a unique identification number. Additionally store metadata attributes of the article.

\paragraph{Language Convention (According to ISO convention)}
\begin{center}
\begin{tabular}{rl}
\hline
S.No. & Language (Convention) \\ \hline
1 & English (\texttt{en}) \\
2 & Hindi (\texttt{hi}) \\
3 & Gujarati (\texttt{gu}) \\
4 & Punjabi (\texttt{pa}) \\
5 & Telugu (\texttt{te}) \\
6 & Kannada (\texttt{kn}) \\
7 & Urdu (\texttt{ur}) \\
8 & Odia (\texttt{od}) \\
9 & Dogri (\texttt{dg}) \\
10 & Kashmiri (\texttt{ks}) \\
11 & Sindhi (\texttt{sd}) \\ \hline
\end{tabular}
\end{center}

\paragraph{File Naming Convention:}
Files will be named using the pattern:
\begin{center}
\texttt{Domain\_Sub-domain\_LanguagePair\_PartNo.xls}
\end{center}

\begin{enumerate}
    \item \textbf{Domain mnemonics:} Three letters will be used for defining domain names e.g. Health: \texttt{HLT}; Governance: \texttt{GOV}; General: \texttt{GEN}.
    \item \textbf{Sub-domain mnemonics:} To be decided.
    \item \textbf{LanguagePair:} Examples are given in the table below.
\end{enumerate}

\begin{center}
\begin{tabular}{rl}
\hline
S.No. & Language (Mnemonics) \\ \hline
1 & English $\rightarrow$ Hindi (\texttt{en-hi}) \\
2 & English $\rightarrow$ Telugu (\texttt{en-te}) \\
3 & Hindi $\leftrightarrow$ Punjabi (\texttt{hi-pa} or \texttt{pa-hi}) \\
4 & Hindi $\leftrightarrow$ Telugu (\texttt{hi-te} or \texttt{te-hi}) \\
5 & Hindi $\leftrightarrow$ Urdu (\texttt{hi-ur} or \texttt{ur-hi}) \\
6 & Hindi $\leftrightarrow$ Gujarati (\texttt{hi-gu} or \texttt{gu-hi}) \\
7 & Hindi $\leftrightarrow$ Kannada (\texttt{hi-kn} or \texttt{kn-hi}) \\
8 & Hindi $\leftrightarrow$ Odia (\texttt{hi-od} or \texttt{od-hi}) \\
9 & Hindi $\leftrightarrow$ Kashmiri (\texttt{hi-ks} or \texttt{ks-hi}) \\
10 & Hindi $\leftrightarrow$ Sindhi (\texttt{hi-sd} or \texttt{sd-hi}) \\
11 & Hindi $\leftrightarrow$ Dogri (\texttt{hi-dg} or \texttt{dg-hi}) \\ \hline
\end{tabular}
\end{center}

\paragraph{PartNo:} Each part of the file can contain a maximum of 200 sentences. For files having more than 200 sentences, they will be split into multiple parts and numbers to files will be assigned accordingly.

\paragraph{Example file names:}
\begin{itemize}
    \item \texttt{HLT\_SBD\_en-hi\_ART1\_01.xls}
    \item \texttt{HLT\_SBD\_en-hi\_ART1\_02.xls}
\end{itemize}

\paragraph{Sample Healthcare sub-domains:}
\begin{center}
\begin{tabular}{rl}
\hline
S. No. & Sub domains (Mnemonics) \\ \hline
1 & Consent Forms (\texttt{CNF}) \\
2 & Awareness material (\texttt{AWN}) \\
3 & Pharma related (\texttt{PHR}) \\ \hline
\end{tabular}
\end{center}

\subsubsection*{d. Meta Information:}
Once the corpora are collected, some corpora related information needs to be included. These are called ``metadata''. The metadata can either be encoded in the corpus text or held in a separate document or database. The following kinds of meta information will be stored:
\begin{itemize}
    \item Title: The title of original linguistic material, such as books, articles, web pages and so on.
    \item Source: Source of the corpora, such as Journal, News Article, Website, Newsletter.
    \item Author/Publisher: The author of the written material.
    \item Creation date: Date of the corpora creation to be mentioned.
    \item Copyright: Shows the composer, publishing company, or the web site of the original linguistic material.
\end{itemize}


\subsubsection*{f. Cleaning and Preprocessing:}
As the majority of the collected corpora is web scraped, we need to preprocess them before they are ready for translation or annotation. The following kinds of preprocessing is required for the monolingual/parallel corpora:
\begin{itemize}
    \item Special/unwanted characters, multiple symbols and repeated punctuation symbols\\
    \textit{Example:} ``Hello!!!, he said ---and went.''
    \item Sentences having invalid characters or sentences which are out of valid Unicode range values defined for a given language (Out of Unicode Boundary Characters).
    \item Duplicate sentences and multi-words not connected with hyphens in the corpus.
    \item Sentences having words containing more than 20 characters. Different words joined due to typing or errors due to non-segmentation e.g. \texttt{hascongratulatedtheIndianCricket}
    \item Very short (approx 3 words) and very long (approx 80 words) sentences.
    \item Data inconsistencies and abbreviations should be mapped to their original form using text normalization techniques.\\
    \textit{Example:}\\
    1) S. No., Sr. No., Sl. No., Sn. No., Serial No.\\
    2) Gender representation by ``male'' and ``female'' and alternates to ``M'' and ``F''.
    \item Bulleted points from the corpora including both ordered and unordered bullets.
    \item Accented characters - Accented characters signify emphasis on a particular word. Words such as r\'{e}sum\'{e}, caf\'{e}, pr\'{o}test, divorc\'{e}, co\"{o}rdinate, expos\'{e}, latt\'{e} etc. These characters need to be converted and standardized into ASCII characters.
    \item Expanding the text contractions for text standardization.\\
    \textit{Example:} don't $\rightarrow$ do not, I'd $\rightarrow$ I would, you're $\rightarrow$ you are.
    \item Words in brackets extraction.
    \item Many symbols with different encodings present in the corpus like:
    \begin{itemize}
        \item four types of dash symbols
        \item different double quotes
        \item Hindi number variations
        \item words having different encodings
    \end{itemize}
\end{itemize}

\subsubsection*{g. Corpora Creation and Annotation Interface}
A centrally monitored interface will facilitate the corpora creation and annotation tasks. Individual tasks will be assigned and trackers will be in place to monitor these activities. This will eventually remove the issues related to formats and file name inconsistencies. The developed corpora would be saved on a server with proper backup facilities.

\subsubsection*{h. Corpora Storage}
All the parallel corpora and annotated corpora will be stored in a central repository. Open source repositories like GitHub/GitLab will be used for this as they also provide the facility of version control. These platforms also provide collaborative features like task management, integration and wikis for documentation.






\end{document}